\documentclass{article} 
\usepackage{iclr2019_conference,times}

\usepackage{amsmath,amsfonts,bm}

\def\eqref#1{equation~\ref{#1}}

\def\1{\bm{1}}

\DeclareMathAlphabet{\mathsfit}{\encodingdefault}{\sfdefault}{m}{sl}
\SetMathAlphabet{\mathsfit}{bold}{\encodingdefault}{\sfdefault}{bx}{n}

\DeclareMathOperator*{\argmin}{arg\,min}

\usepackage{hyperref}
\usepackage{url}
\usepackage{graphicx}
\usepackage{subcaption,enumitem}
\usepackage{booktabs}
\usepackage{footmisc, bm}

\title{Interpolation-Prediction Networks for \\ Irregularly Sampled Time Series}

\author{Satya Narayan Shukla \\
College of Information and Computer Sciences\\
University of Massachusetts Amherst\\
\texttt{snshukla@cs.umass.edu} \\
\AND
Benjamin M. Marlin \\
College of Information and Computer Sciences\\
University of Massachusetts Amherst\\
\texttt{marlin@cs.umass.edu} \\
}

\newcommand{\cut}[1]{}
\newcommand{\mbf}[1]{\mathbf{#1}}

\iclrfinalcopy 

\begin{document}

\maketitle

\begin{abstract}
\cut{
While the volume of electronic health records (EHR) data continues to grow,
it remains rare for hospital systems to capture dense physiological
data streams, even in the data-rich intensive care unit setting. 
Instead, typical EHR records
consist of sparse and irregularly observed multivariate time series,
which are well understood to present particularly challenging problems
for machine learning methods. In this paper, we present a new deep learning
architecture for addressing this problem based on the use of 
a semi-parametric interpolation network followed by the 
application of a prediction network. 
The interpolation network allows for information to be shared across multiple
dimensions during the interpolation stage, while any standard
deep learning model can be used for the prediction network. 
We investigate the performance
of several versions of this architecture on the problems of mortality
and length of stay prediction. }

In this paper, we present a new deep learning architecture for addressing the problem of supervised learning with sparse and irregularly sampled multivariate time series. The architecture is based on the use of a semi-parametric interpolation network followed by the application of a prediction network. The interpolation network allows for information to be shared across multiple dimensions of a multivariate time series during the interpolation stage, while any standard deep learning model can be used for the prediction network. This work is motivated by the analysis of physiological time series data in electronic health records, which are sparse, irregularly sampled, and multivariate. We investigate the performance of this architecture on both classification and regression tasks, showing that our approach outperforms a range of baseline and recently proposed models.\footnote{
Our implementation is available at : \url{https://github.com/mlds-lab/interp-net}}

\end{abstract}

\section{Introduction}

Over the last several years, there has been significant progress 
in developing specialized models and architectures that can
accommodate sparse and irregularly sampled time series
as input \citep{marlin-ihi2012, li2015classification, li2016scalable, lipton2016directly,  futoma2017improved, che2016recurrent}. 
An irregularly sampled time series 
is a sequence of samples with irregular intervals between their observation times.
Irregularly sampled data are considered to be sparse when the intervals
between successive observations are often large.
Of particular interest in the supervised learning setting are 
methods that perform end-to-end learning directly using 
multivariate sparse and irregularly sampled time series as input without
the need for a separate interpolation or imputation step. 

In this work, we present a new model architecture for
supervised learning with multivariate sparse and irregularly sampled data: 
Interpolation-Prediction Networks. 
The architecture is based on the use of several semi-parametric interpolation layers
organized into an interpolation network, followed by the 
application of a prediction network that can leverage any standard deep learning model.
In this work, we use GRU networks \citep{gru} as the prediction
network.

The interpolation network allows for information contained in each input time series
to contribute to the interpolation of all other time series
in the model. The parameters of the interpolation and prediction networks
are learned end-to-end via a composite objective function consisting 
of supervised and unsupervised components. The interpolation network serves 
the same purpose as the multivariate Gaussian process used in 
the work of \citet{futoma2017improved}, but remove the restrictions 
associated with the need for a positive definite covariance matrix. 

Our approach also allows us to compute an explicit multi-timescale representation
of the input time series, which we use to isolate information
about transients (short duration events) from broader trends.  
Similar to the work of \citet{lipton2016directly} and  \citet{che2016recurrent},
our architecture also explicitly leverages a separate information
channel related to patterns of observation times. However, our
representation uses a semi-parametric intensity function representation
of this information that is more closely related to the work of \citet{lasko2014efficient}
on modeling medical event point processes.

Our architecture thus produces three 
output time series for each input time series: a smooth
interpolation modeling broad trends in the input, a short 
time-scale interpolation modeling transients, and an intensity
function modeling local observation frequencies. 

This work is motivated by problems in the analysis of electronic health records (EHRs)
\citep{marlin-ihi2012, lipton2016directly, futoma2017improved, che2016recurrent}.
It remains rare for hospital systems to capture dense physiological
data streams. Instead, it is common for the physiological time series data in electronic health records
to be both sparse and irregularly sampled.
The additional issue of the lack of alignment in the observation times
across physiological variables is also very common.

We evaluate the proposed architecture on two datasets for both classification and regression tasks.
Our approach outperforms a variety of simple baseline
models as well as the basic and advanced GRU models introduced by 
\citet{che2016recurrent} across several metrics. We also compare our model with to the Gaussian process adapter \citep{li2016scalable} and multi-task Gaussian process RNN classifier \citep{futoma2017improved}. 
Further, we perform full ablation testing of the information channels 
our architecture can produce to assess their impact on classification
and regression performance. 

\section{Related Work}
\label{sec:related}

The problem of interest in this work is learning supervised 
machine learning models from sparse and irregularly sampled
multivariate time series. As described in the introduction,
a sparse and irregularly sampled time series is a sequence of 
samples with large and irregular intervals between their 
observation times. 

Such data commonly occur in electronic 
health records, where they can represent a significant problem 
for both supervised and unsupervised learning methods \citep{yadav2018mining}.
Sparse and irregularly sampled time series data also 
occur in a range of other areas with similarly complex
observation processes including
climate science \citep{schulz1997spectrum},
ecology \citep{clark2004population},
biology \citep{ruf1999lomb},
and astronomy \citep{scargle-astro1982}. 

A closely related (but distinct) problem is performing supervised learning in the
presence of missing data \citep{batista2003analysis}. The primary difference is that
the missing data problem is generally defined with respect to
a fixed-dimensional feature space \citep{little2014statistical}. In the irregularly 
sampled time series problem, observations typically 
occur in continuous time and there may be no notion of 
a ``normal" or ``expected" sampling frequency for some
domains. 

Methods for dealing with missing data in supervised learning 
include the pre-application of imputation methods
\citep{sterne2009multiple}, and learning 
joint models of features and labels \citep{williams2005incomplete}. Joint models can either 
be learned generatively to optimize the joint likelihood of 
features and labels, or discriminately to optimize the 
conditional likelihood of the labels. The problem of irregular
sampling can be converted to a missing data problem by discretizing 
the time axis into non-overlapping intervals. Intervals with no
observations are then said to contain missing values. 

This is the
approach taken to deal with irregular sampling by \cite{marlin-ihi2012}
as well as \cite{lipton2016directly}. This
approach forces a choice of discretization interval length.
When the intervals are long, there will be less missing data, but
there can also be multiple observations in the same interval, which
must be accounted for using ad-hoc methods. When the intervals are shorter,
most intervals will contain at most one value, but many intervals 
may be empty. Learning is generally harder as the amount of missing data
increases, so choosing a discretization interval length must be 
dealt with as a hyper-parameter of such a method.

One important feature of missing data problems is the potential for
the sequence of observation times to itself be informative
\citep{little2014statistical}. Since the set of missing data indicators 
is always observed, this information is typically easy to condition
on. This technique has been used successfully to improve models
in the domain of recommender systems \citep{salakhutdinov2007restricted}. 
It was also used by \cite{lipton2016directly} to improve performance
of their GRU model.

The alternative to pre-discretizing irregularly sampled time series to
convert the problem of irregular sampling into the problem of missing data
is to construct models with the ability to directly use
an irregularly sampled time series as input. The machine learning
and statistics literature include several models with this ability.
In the probabilistic setting, Gaussian process models have the ability
to represent continuous time data via the use of mean and covariance
functions \citep{rasmussen2006gaussian}. These models have non-probabilistic
analogues that are similarly defined in terms of kernels. 

For example, \cite{Lu2008} present a kernel-based method that can be used
to produce a similarity function between two irregularly sampled time
series. \cite{li2015classification} subsequently provided a  
generalization of this approach to the case of kernels between Gaussian
process models. \cite{li2016scalable} showed how the re-parameterization trick
\citep{kingma2015variational} could be used to extend these ideas to enable 
end-to-end training of a deep neural network model
(feed-forward, convolutional, or recurrent)
stacked on top of a Gaussian process layer. While the basic model of \cite{li2016scalable}
was only applied to univariate time series, in follow-up work the 
model was extended to multivariate time series using a multi-output
Gaussian process regression model \citep{futoma2017improved}. 
However, modeling multivariate time series within this framework 
is quite challenging due to the constraints on the covariance function
used in the GP layer. \citet{futoma2017improved}
deal with this problem using a sum of separable kernel
functions \citep{bonilla2008multi}, which limit the expressiveness
of the model.

An important property of the above models is that they allow for incorporating 
all of the information from all available time points into a global interpolation 
model. Variants differ in terms of whether they only leverage the posterior
mean when the final supervised problem is solved, or whether the whole posterior is 
used. A separate line of work has looked at the use of more local interpolation
methods while still operating directly over continuous time inputs.

For example, \citet{che2016recurrent}
presented several methods based on  gated recurrent unit (GRU) networks
\citep{gru} combined with simple imputation methods
including mean imputation and forward filling with past values. 
\citet{che2016recurrent} additionally considered an approach 
that takes as input a sequence consisting of both the
observed values and the timestamps at which those values were observed. 
The previously observed input value is decayed over time 
toward the overall mean. In another variant the hidden states are 
similarly decayed toward zero. \cite{Yoon_mRNN} presented another similar approach based on multi-directional RNN which operate across streams in addition to
within streams. However, these models are limited to using 
global information about the structure of the time series via its 
empirical mean value, and current or past information about
observed values. The global structure of the time series is not directly
taken into account.  

\cite{che18a} focus on a similar problem of modeling multi-rate multivariate time series data.  This is similar to the problem of interest in that the observations across time series can be unaligned. The difference is that the observations in each time series are uniformly spaced, which is a simpler case. In the case of missing data, they use forward or linear interpolation, which again does not capture the global structure of time series.  Similarly, \cite{auto_cnn} presented an autoregressive framework for regression tasks with irregularly sampled time series data. It is not clear how it can be extended for classification. 

The model proposed in this work is similar to that of \cite{li2016scalable}
and \cite{futoma2017improved} in the sense that it consists of global
interpolation layers. The primary difference is that these prior approaches 
used Gaussian process representations within the interpolation layers.
The resulting computations can be expensive and, as noted, the design of 
covariance functions in the multivariate case can be challenging. By contrast,
our proposed model uses semi-parametric, deterministic, feed-forward interpolation
layers. These layers do not encode uncertainty, but they do allow for
very flexible interpolation both within and across layers. 

Also similar to \cite{li2016scalable} and \cite{futoma2017improved}, the interpolation
layers in our architecture produce regularly sampled interpolants that can 
serve as inputs for arbitrary, unmodified, deep classification and
regression networks. This is in contrast to the approach of 
\citet{che2016recurrent}, where a recurrent network architecture was directly 
modified, reducing the modularity of the approach. 
Finally, similar to \cite{lipton2016directly}, our model includes information
about the times at which observations occur. However, instead of pre-discretizing
the inputs and viewing this information in terms of a binary observation mask
or set of missing data indicators, we directly model the sequence of
observation events as a point process in continuous time 
using a semi-parametric intensity function
\citep{lasko2014efficient}.


\section{Model Framework}

In this section, we present the proposed modeling framework. We begin by presenting notation, 
followed by the model architecture and learning criteria.

\subsection{Notation}

We let $\mathcal{D}=\{(\mbf{s}_n,y_n) | n=1,...,N\}$ represent a data set containing
$N$ data cases. An individual data case consists of
a single target value $y_n$ (discrete for classification
and real-valued in the case of regression), as well as a $D$-dimensional, sparse and irregularly
sampled multivariate time series $\mbf{s}_n$. Different dimensions $d$ of the multivariate
time series can have observations at different times, as well as different total
numbers of observations $L_{dn}$. Thus, we represent time series $d$ for data case $n$ as a tuple
$\mbf{s}_{dn}=(\mbf{t}_{dn}, \mbf{x}_{dn})$  where $\mbf{t}_{dn}=[t_{1dn},...,t_{L_{dn}dn}]$
is the list of time points at which observations are defined and 
$\mbf{x}_{dn}=[x_{1dn},...,x_{L_{dn}dn}]$ is the corresponding list of observed values.

\subsection{Model Architecture}

The overall model architecture consists of two main components: an interpolation
network and a prediction network. The interpolation network interpolates
the multivariate, sparse, and irregularly sampled input time series against 
a set of reference time points $\mbf{r}=[r_1,...,r_T]$. We assume that all of the 
time series are defined within a common time interval (for example, the first 24 or 48  
hours after admission for MIMIC-III dataset). The $T$ reference time points $r_t$ are chosen to be evenly 
spaced within that interval. In this work, we propose a two-layer interpolation network with each layer performing a different type of interpolation.

\begin{figure}[t]
\includegraphics[width=\linewidth]{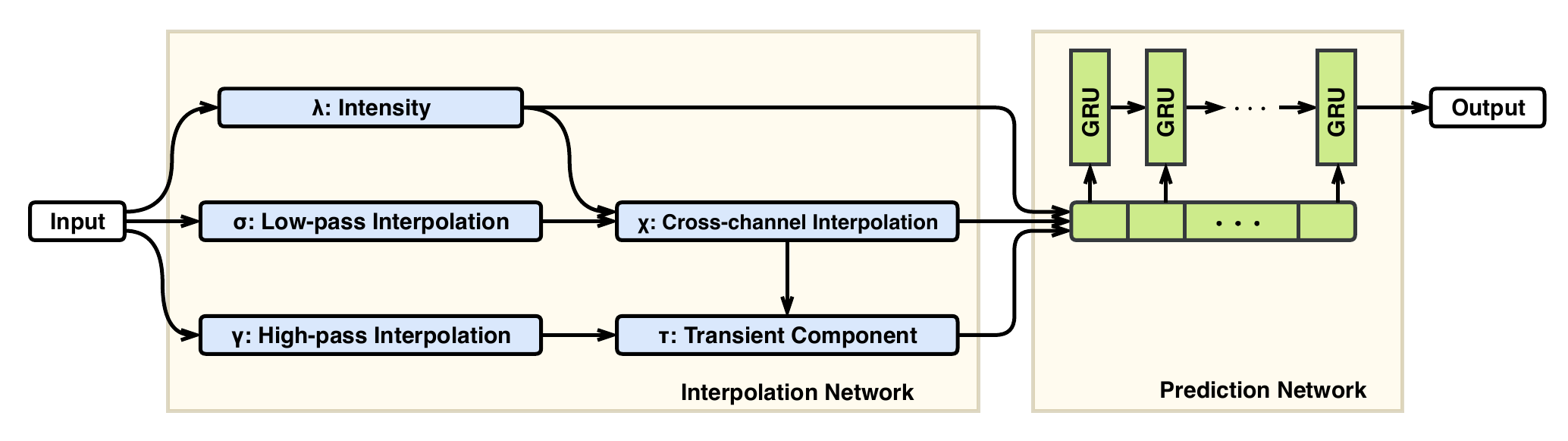}
\centering
\caption{Architecture of the proposed model}
\label{fig:model}
\end{figure}

The second component, the prediction network, takes the output of the interpolation
network as its input and produces a prediction $\hat{y}_n$ for the target variable.
The prediction network can consist of any standard supervised 
neural network architecture (fully-connected feedforward, convolutional, recurrent, etc).
Thus, the architecture is fully modular with respect to the use of different prediction
networks. In order to train the interpolation network, the architecture also includes
an auto-encoding component to provide an unsupervised learning signal in addition to
the supervised learning signal from the prediction network. Figure \ref{fig:model} shows the architecture of the proposed model. We describe the components
of the model in detail below. 

\subsubsection{Interpolation Network} We begin by describing the interpolation network. 
The goal of the interpolation network is to provide a collection of interpolants 
of each of the $D$ dimensions of an input multivariate time series defined at the 
$T$ reference time points $\mbf{r}=[r_1,...,r_T]$. In this work, we use a total
of $C=3$ outputs for each of the 
$D$ input time series. The three outputs (discussed in detail below) capture smooth trends, transients, and observation intensity information. We define $f_{\theta}(\mbf{r},\mbf{s}_n)$ to be the function computing the output $\hat{\mathbf{s}}_n$ of the interpolation network. The output $\hat{\mathbf{s}}_n$ is a fixed-sized array with dimensions $(DC)\times T$ for all inputs $\mbf{s}_n$.

The first layer in the interpolation network separately performs three semi-parametric 
univariate transformations for each of the $D$ time series. Each transformation
is based on a radial basis function (RBF) network to accommodate continuous time observations. The transformations are
a low-pass (or smooth) interpolation $\bm{\sigma}_{d}$, a high-pass (or non-smooth) interpolation $\bm{\gamma}_{d}$ and an intensity function $\bm{\lambda}_{d}$. These transformations are computed at reference time point $r_k$ for each data case and each input time series $d$ as shown in Equations \ref{eq:interp0}, \ref{eq:interp1}, \ref{eq:interp2} and \ref{eq:interp3}.\footnote{We drop the data case index $n$ for brevity in the equations below.} The smooth interpolation $\bm{\sigma}_{d}$ uses a squared exponential kernel with  parameter $\alpha_d$, while the non-smooth interpolation $\bm{\gamma}_{d}$ uses a squared exponential kernel with parameter
$\kappa \alpha_d$ for $\kappa>1$. 

\vspace{-1mm}
\begin{align}
\label{eq:interp0}
Z(r, \mbf{t}, \alpha) &= \sum_{t \in \mbf{t}} w(r,t, \alpha)  \:, \:\:\:\:\: w(r,t,\alpha) = \exp(-\alpha(r-t)^2)\\	
 \label{eq:interp1}
\lambda_{kd} &=  h_\theta^\lambda(r_k, \mbf{t}_d, \mbf{x}_d) = Z(r_k, \mbf{t}_d, \alpha_d)\\ 
%
 \label{eq:interp2}
\sigma_{kd} &= h_\theta^\sigma(r_k, \mbf{t}_d, \mbf{x}_d) = \frac{1}{Z(r_k, \mbf{t}_d, \alpha_d)}  {\displaystyle \sum_{j=1}^{L_{dn}} w(r_k,t_{jd}, \alpha
_d) \:x_{jd}} \\
 \label{eq:interp3}
\gamma_{kd} &= h_\theta^\gamma(r_k, \mbf{t}_d, \mbf{x}_d) = \frac{1}{Z(r_k, \mbf{t}_d,\kappa \alpha_d)}  {\displaystyle \sum_{j=1}^{L_{dn}} w(r_k,t_{jd}, \kappa\alpha_d) \:x_{jd}} 
\end{align}

The second interpolation layer merges information across all $D$
time series at each reference time point by taking into account
learnable correlations $\rho_{dd'}$ across all time series. This
results in a cross-dimension interpolation $\bm{\chi}_{d}$ for 
each input dimension $d$. We further define a transient component  
$\bm{\tau}_{d}$ for each input dimension $d$ as the difference between
the high-pass (or non-smooth) interpolation $\bm{\gamma}_{d}$  from the first layer and the
smooth cross-dimension interpolation  $\bm{\chi}_{d}$, as shown in Equation \ref{eq:interp4}. 

\vspace{-1mm}
\begin{align}
 \label{eq:interp4}
 \chi_{kd} = h_\theta^\chi(r_k, \mbf{s})=\frac{\sum_{d'} \rho_{dd'} \: \lambda_{kd'} \: \sigma_{kd'}}{\sum_{d'} \lambda_{kd'}} \: , &&
 \tau_{kd} = h_\theta^\tau(r_k, \mbf{s}) =  \gamma_{kd} - \chi_{kd}
\end{align}

In the experiments presented in the next section, we use a total of three
interpolation network outputs per dimension $d$ as the input to the prediction network.
We use the smooth, cross-channel interpolants $\bm{\chi}_d$ to capture
smooth trends, the transient components $\bm{\tau}_d$ to capture transients, and the 
intensity functions $\bm{\lambda}_d$ to capture information about where observations occur in time.

\subsubsection{Prediction Network} Following the application of the 
interpolation network, all $D$ dimensions of the input multivariate time series have been
re-represented in terms of $C$ outputs defined on the regularly spaced 
set of reference time points $r_1,...,r_T$ (in our
experiments, we use $C=3$ as described above).
Again, we refer to the complete set of interpolation network outputs 
as $\hat{\mathbf{s}}_{n}=f_{\theta}(\mbf{r},\mbf{s}_n)$, which can be represented as a matrix
of size $(DC)\times T$. 

The prediction network must take $\hat{\mathbf{s}}_{n}$ as input and
output a prediction $\hat{y}_n=g_{\phi}(\hat{\mbf{s}}_n)=
g_{\phi}(f_{\theta}(\mbf{r},\mbf{s}_n))$ of the target value $y_n$
for data case $n$. There are many possible choices for this
component of the model. For example, the matrix $\hat{\mathbf{s}}_{n}$ 
can be converted into a single long vector and provided as 
input to a standard multi-layer feedforward network. 
A temporal convolutional model or a recurrent model like a 
GRU or LSTM can instead be applied to time slices of the 
matrix $\hat{\mathbf{s}}_{n}$. In this work, we conduct experiments
leveraging a GRU network as the prediction network.

\subsubsection{Learning} To learn the model parameters, we use a 
composite objective function consisting of a supervised component
and an unsupervised component. This is due to the fact that the 
supervised component alone is insufficient to learn reasonable
parameters for the interpolation network parameters given the amount
of available training data. The unsupervised component used 
corresponds to an autoencoder-like loss function. However,
the semi-parametric RBF interpolation layers have the ability to
exactly fit the input points by setting the RBF kernel parameters
to very large values. 

To avoid this solution and force the interpolation
layers to learn to properly interpolate the input data, it is necessary
to hold out some observed data points $x_{jdn}$ during learning and then
to compute the reconstruction loss only for these data points. This is a 
well-known problem with high-capacity autoencoders, and past work 
has used similar strategies to avoid the problem of trivially 
memorizing the input data without learning useful structure.

To implement the autoencoder component of the loss, we introduce
a set of masking variables $m_{jdn}$ for each data point $(t_{jdn}, x_{jdn})$.
If $m_{jdn}=1$, then we remove the data point $(t_{jdn}, x_{jdn})$ as an input
to the interpolation network, and include the predicted value of this time point
when assessing the autoencoder loss. We use the shorthand notation $\mbf{m}_n \odot \mbf{s}_n$ to represent
the subset of values of $\mbf{s}_n$ that are masked out, and $(1-\mbf{m}_n) \odot \mbf{s}_n$
to represent the subset of values of $\mbf{s}_n$ that are not masked out. 
The value $\hat{x}_{jdn}$ that we predict for a masked input at time  
point $t_{jdn}$ is the value of the smooth cross-channel interpolant at that time point, calculated
based on the un-masked input values: $\hat{x}_{jdn} = h_\theta^{\chi}(t_{jdn}, (1-\mbf{m}_n) \odot \mbf{s}_n)$.

We can now define the learning objective for
the proposed framework. We let $\ell_P$ be the loss for the prediction
network (we use cross-entropy loss for classification and squared 
error for regression). We let $\ell_I$ be the interpolation network autoencoder loss (we use standard squared error).
We also include $\ell_2$ regularizers for both the interpolation
and prediction networks parameters. $\delta_I$, $\delta_P$, and $\delta_R$ are
hyper-parameters that control the trade-off between the components of the
objective function.
\begin{align}
	\theta_*,\phi_* &=\argmin_{\theta,\phi} \sum_{n=1}^N \ell_P(y_n, g_{\phi}(f_{\theta}(\mbf{s}_n))
	+ \delta_I \Vert \theta \Vert_2^2 + \delta_P \Vert \phi \Vert_2^2 \\
	\nonumber &\;\;\;\;\;\;\;\;\;\;\;\;+ \delta_R \sum_{n=1}^N \sum_{d=1}^D \sum_{j=1}^{L_{dn}} m_{jdn} \ell_I(x_{jdn},h_\theta^\chi(t_{jdn}, (1-\mbf{m}_n) \odot \mbf{s}_n))
\end{align}


\section{Experiments and Results}

In  this section, we present experiments based on  both classification and regression tasks with sparse and irregularly sampled multivariate time series.
In both cases, the input to the prediction network is a sparse and irregularly sampled time series,
and the output is a single scalar representing either the predicted class or the regression target variable.
We test the model framework on two publicly available real-world datasets: MIMIC-III \footnote{MIMIC-III  is  available at \url{https://mimic.physionet.org/}} $-$ a multivariate time series dataset consisting of sparse and irregularly sampled physiological signals collected at Beth Israel Deaconess Medical Center
from 2001 to 2012 \citep{johnson2016mimic}, and UWaveGesture \footnote{UWaveGestureLibraryAll is available at \url{http://timeseriesclassification.com}.} $-$ a univariate time series data set consisting of simple gesture patterns divided into eight categories \citep{uwave}. Details of each dataset can be found in the Appendix \ref{dataset}.  We use the MIMIC-III mortality and length of stay prediction tasks as example classification and regression tasks with multivariate time series. We use the UWave gesture  classification task for assessing training time and performance relative to univariate baseline models.


\subsection{Baseline Models}
We compare our proposed model to a number of baseline approaches including off-the-shelf classification
and regression models learned using basic features, as well as more recent approaches
based on customized neural network models.

\subsubsection{Non-Neural Network Baselines}
For non-neural network baselines, we evaluate Logistic Regression \citep{hosmer2013applied}, Support Vector Machines (SVM) \citep{cortes1995}, Random Forests (RF) \citep{breiman_rf} and AdaBoost \citep{adaboost} for the classification task.

For the length of stay prediction task, we apply Linear Regression \citep{hastie01statisticallearning}, Support Vector Regression (SVR), AdaBoost Regression \citep{drucker} and Random Forest Regression. Standard instances of all of these models require fixed-size feature representations. We use temporal discretization with forward filling to create fixed-size representation in case of missing data and use this representation as feature set for non-neural network baselines. 

\subsubsection{Neural Network Models}
We compare to several existing deep learning baselines built on GRUs using simple interpolation or imputation approaches. In addition, we compare to current state-of-the-art models for mortality prediction including the work of \cite{che2016recurrent}. Their work proposed to handle irregularly sampled and missing data using recurrent neural networks (RNNs) by introducing temporal decays in the input and/or hidden layers.  We also evaluate the scalable end-to-end Gaussian process adapter  \citep{li2016scalable} as well as multi-task Gaussian process RNN classifier \citep{futoma2017improved} for
irregularly sampled univariate and multivariate time series classification respectively.  This work is discussed in detail in Section 
\ref{sec:related}. The complete set of models that we compare to is as follows:

\begin{itemize}
\item{\bf GP-GRU:} End-to-end Gaussian process with GRU as classifier.
\item {\bf GRU-M:} Missing observations replaced with the global mean of the variable across the training examples. 
\item {\bf GRU-F:} Missing values set to last observed measurement within that time series (referred to as forward filling). 
\item {\bf GRU-S:}  Missing values replaced with the global mean. Input is concatenated with masking variable
and time interval indicating how long the particular variable is missing. 
\item {\bf GRU-D:}  In order to capture richer information, decay is introduced in the input as well as hidden layer of a GRU. Instead of replacing missing values with the last measurement, missing values are decayed over time towards the empirical mean. 
\item{\bf GRU-HD:} A variation of GRU-D where decay in only introduced in the hidden layer.

\end{itemize}
\subsection{Results}
In this section, we present the results of the classification and regression
experiments, as well as the results of ablation testing of the internal structure of the 
interpolation network for the proposed model. We use the UWaveGesture dataset to assess the training time and classification performance relative to the baseline models. We use the standard train and test sets (details are given in appendix \ref{dataset}). We report the training time taken for convergence along  with accuracy on test set. 

For MIMIC-III, we create our own dataset (appendix \ref{dataset}) and report the results of a 5-fold cross validation experiment in terms of the average area under the ROC curve (AUC score), average area under the precision-recall curve (AUPRC score), and average cross-entropy loss for the classification task. For  the regression task, we use average median absolute error and average fraction of explained variation  (EV) as metrics. We also report the standard deviation over cross validation folds for all metrics. 

\begin{figure}[h]
\centering
  \includegraphics[width=0.6\linewidth]{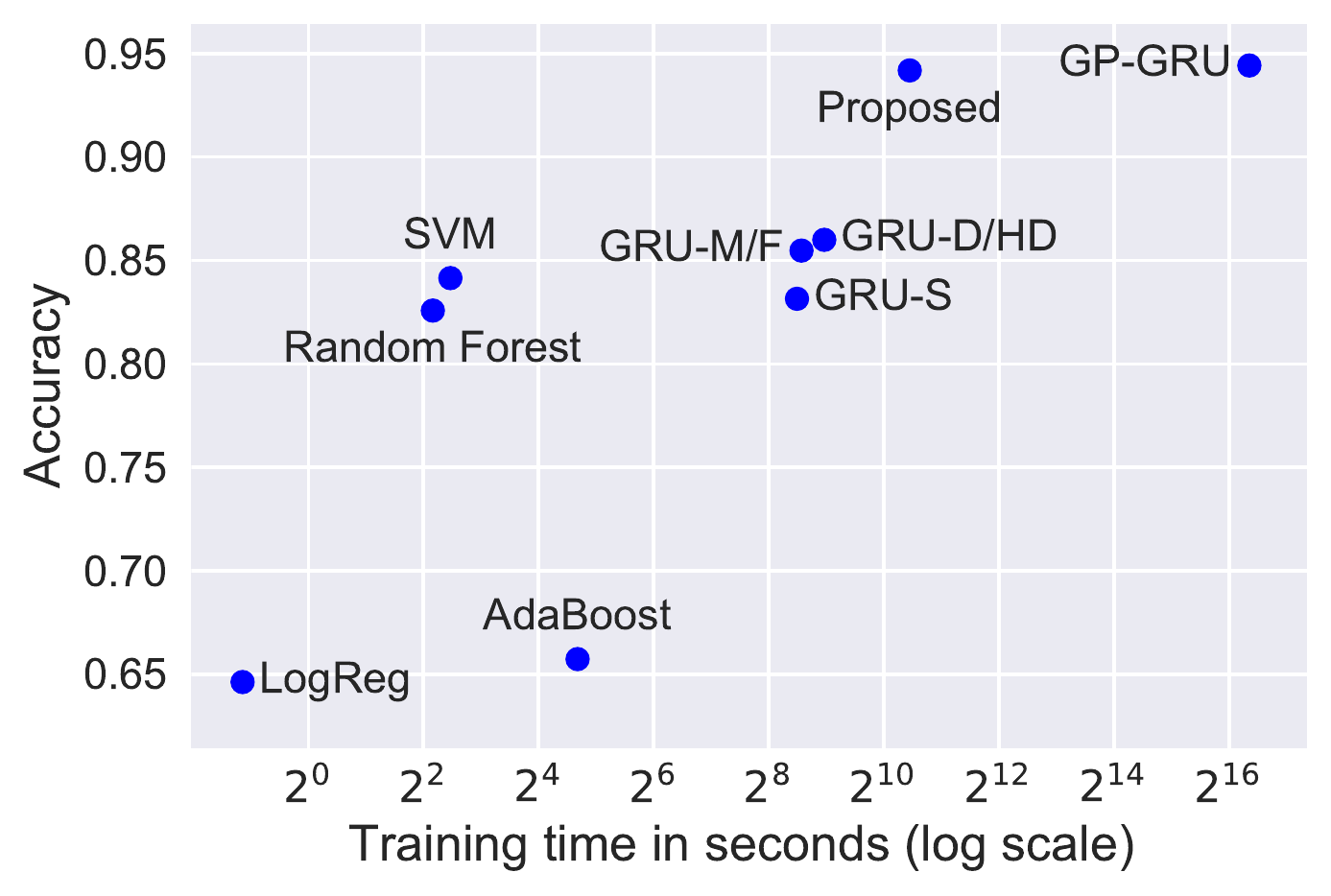}
  \captionof{figure}{Classification performance on the UWaveGesture dataset. Models with almost same performance are shown with the same dot e.g. (GRU-M, GRU-F ) and (GRU-D, GRU-HD).}
  \label{fig:uwave}
\end{figure}

\cut{
\begin{table}[h!]
\caption{Classification performance on UWaveGesture dataset (all time measures are in seconds)}
\footnotesize
\begin{center}
\begin{tabular}{ l c c c c} 
 \toprule
 {\bf Model} & {\bf Accuracy}  &  {\bf Time per epoch} & {\bf  \# Iterations } & {\bf Training time} \\
 \midrule
 
 {LogReg} & $0.6462$ & $4.532 \times 10^{-3}$ &$100$ & $4.532 \times 10^{-1}$\\
  {SVM} &  $0.8415$ & $2.769 \times 10^{-2}$ & $200$ & $5.538  \times 10^0$\\
  AdaBoost & $0.6573$ &$-$ & $-$ & $2.562 \times 10^1$ \\ 
 
  RF & $0.8258$ & $-$ & $-$ & $4.483  \times 10^0$\\
 
  GRU-M & $0.6640$ & 	$0.968 \times 10^0$& $500$ & $4.840 \times 10^2$ \\ 
 
 GRU-F  & $0.6640$ & 	$0.968  \times 10^0$& $500$ & $4.840\times 10^2$ \\
 
  GRU-S & $0.7355$ & 	$1.115  \times 10^0$ & $250$ & $2.787 \times 10^2$ \\ 

 GRU-D & $0.6194$ & 	$1.362  \times 10^0$&	$500$ & $6.810 \times 10^2$ \\
 
 GRU-HD & $0.6149$ & $1.380  \times 10^0$ & $500$ & $6.900 \times 10^2$   \\ 
 GP-GRU & $\bf{0.9251}$ & $1.673 \times 10^3$ &$50$ & $  { 8.365 \times 10^4}$ \\
 
 {\bf Proposed} &  $\bf{0.9230}$ &	$4.875  \times 10^0$&	 $300$  & $ 1.463\times 10^3 $\\ 
\bottomrule
 \end{tabular}
\end{center}
\label{table:uwave}
\end{table} 
}

\begin{table}[h]
\caption{Performance on Mortality (classification) and Length of stay prediction (regression) tasks on MIMIC-III. Loss: Cross-Entropy Loss, MedAE: Median Absolute Error (in days), EV: Explained variance}
\footnotesize
\begin{center}
\begin{tabular}{ l c c c c c} 
 \toprule
 {\bf Model}&  \multicolumn{3}{c}{\bf Classification} &  \multicolumn{2}{c}{\bf Regression}\\
 \midrule
 {} & {\bf AUC} & {\bf AUPRC} & {\bf Loss} & {\bf MedAE} & {\bf EV score}\\
  \midrule
 {Log/LinReg} & $0.772\pm 0.013$ & $0.303\pm0.018$  & $0.240\pm 0.003$  & $3.528\pm0.072$&	$0.043\pm0.012$\\ 
  {SVM} & $0.671\pm 0.005$ & $0.300\pm0.011$& $0.260\pm 0.002$&	$3.523\pm0.071$&	$0.042\pm0.011$\\ 
 
  AdaBoost & $0.829\pm0.007$& $0.345\pm0.007$	&$0.663\pm0.000$ &	$4.517\pm0.234$& 	$0.100\pm0.012$\\ 
 
  RF & $0.826\pm 0.008$ &	$0.356\pm0.010$&	$0.315\pm0.025$ &	$3.113\pm0.125$&	$0.117\pm0.035$  \\ 
 
  GRU-M & $0.831\pm0.007$ & 	$0.376\pm0.022$&	$0.220\pm0.004$ &	$3.140\pm0.196$&	$0.131\pm0.044$\\ 
 
 GRU-F & $0.821\pm0.007$ &  $0.360\pm0.013$&		 $0.224\pm0.003$&	$3.064\pm0.247$& 	$0.126\pm0.025$ \\
 
  GRU-S & $0.843\pm0.007$ & 	$0.376\pm0.014$ &		$0.218\pm0.005$  &	$2.900\pm0.129$&	$0.161\pm0.025$\\ 

 GRU-D & $0.835\pm0.013$ & 	$0.359\pm0.025$&	$0.225\pm0.009$ &  ${\bf 2.891\pm0.103}$&	$0.146\pm0.051$\\
 GRU-HD & $0.845\pm0.006$ & $0.390\pm0.010$ & $0.215\pm0.004$ & ${\bf 2.893\pm0.155}$ &	$0.158\pm0.037$ \\ 
GP-GRU & $0.847 \pm 0.007$ & $0.377 \pm 0.017$ & $0.215 \pm 0.004$ & ${\bf 2.847 \pm 0.079}$ &  $0.217 \pm 0.020$\\
 {\bf Proposed } &  ${\bf 0.853\pm 0.007 }$ &	${\bf 0.418\pm0.022}$&	 ${\bf 0.210 \pm 0.004}$  & ${\bf 2.862\pm0.166}$ & ${\bf 0.245\pm0.019}$\\ 
\bottomrule
 \end{tabular}
\end{center}
\label{table:1}
\end{table}

\cut{
\begin{table}[t]
\caption{Classification performance for the mortality prediction task}
\begin{center}
\begin{tabular}{ l c c c} 
 \toprule
 {\bf Models} & {\bf AUC score}  &  {\bf AUPRC Score} & {\bf Cross-Entropy Loss } \\
 \midrule
 {Logistic Regression} & $0.772\pm 0.013$ & $0.303\pm0.018$  & $0.240\pm 0.003$   \\ 
  {SVM} & $0.671\pm 0.005$ & $0.300\pm0.011$& $0.260\pm 0.002$\\ 
 
  AdaBoost & $0.829\pm0.007$& $0.345\pm0.007$	&$0.663\pm0.000$ \\ 
 
  Random Forest & $0.826\pm 0.008$ &	$0.356\pm0.010$&	$0.315\pm0.025$  \\ 
 
  GRU Mean & $0.831\pm0.007$ & 	$0.376\pm0.022$&	$0.220\pm0.004$ \\ 
 
 GRU Forward & $0.821\pm0.007$ &  $0.360\pm0.013$&		 $0.224\pm0.003$ \\
 
  GRU Simple & $0.843\pm0.007$ & 	$0.376\pm0.014$ &		$0.218\pm0.005$ \\ 

  GRU Hidden Decay & $0.845\pm0.006$ & $0.390\pm0.010$ & $0.215\pm0.004$  \\ 

 GRU-D & $0.835\pm0.013$ & 	$0.359\pm0.025$&	$0.225\pm0.009$ \\

 {\bf Proposed Model} &  ${\bf 0.853\pm 0.007 }$ &	${\bf 0.418\pm0.022}$&	 ${\bf 0.210 \pm 0.004}$ \\ 
\bottomrule
 \end{tabular}
\end{center}
\label{table:1}
\end{table} 
}

Training and implementation details can be found in appendix \ref{implementation}. Figure \ref{fig:uwave} shows the classification performance on the UWaveGesture dataset. The proposed model and the Gaussian process adapter \citep{li2016scalable} significantly outperform the rest of the baselines. However, the proposed model achieves similar performance to the Gaussian process adapter, but with a 50x speed up (note the log scale on the training time axis). On the other hand, the training time of the proposed model is approximately  the same order as other GRU-based models, but it achieves much better accuracy. 
 
Table \ref{table:1} compares the predictive performance of the mortality and length of stay prediction task on MIMIC-III. We note that in highly skewed datasets as is the case of MIMIC-III, AUPRC  \citep{auprc} can give better insights about the classification performance as compared to AUC score.
The proposed model consistently achieves the best average score over all the metrics. 
We note that a paired t-test indicates that the proposed model results in statistically significant improvements over all baseline models $(p<0.01)$ with respect to all the metrics except median absolute error. 
The version of the proposed model used in this experiment includes all three interpolation network
outputs (smooth interpolation, transients, and intensity function).

An ablation study shows that the results on the regression task can be further improved by using only two outputs (transients, and intensity function), achieving statistically significant improvements over all the baselines. Results for the ablation study are given in Appendix \ref{ablation}. Finally, we compare the proposed model with multiple baselines on a previous MIMIC-III benchmark dataset \citep{benchmark}, which uses a reduced number of cohorts as compared to the one used in our experiments. Appendix \ref{benchmark} shows the results on this benchmark dataset, where our proposed approach again outperforms prior approaches.

\section{Discussion and Conclusions}

In this paper, we have presented a new framework for dealing with the
problem of supervised learning in the presence of sparse
and irregularly sampled time series. The proposed framework is fully
modular. It uses an interpolation network to accommodate the complexity
that results from using sparse and irregularly sampled data as
supervised learning inputs, followed by the application of 
a prediction network that operates over the regularly spaced and fully
observed, multi-channel output provided by the interpolation network.
The proposed approach also addresses some difficulties with prior 
approaches including the complexity of the Gaussian process
interpolation layers used in \citep{li2016scalable, futoma2017improved}, and 
the lack of modularity in the approach of \cite{che2016recurrent}.
Our framework also introduces novel elements including the
use of semi-parametric, feed-forward interpolation layers,
and the decomposition of an irregularly sampled 
input time series into multiple distinct information channels.
Our results show statistically significant improvements for both classification and regression tasks
 over a range of
baseline and state-of-the-art methods.

\section*{Acknowledgements}

This work was supported by the National Science Foundation under Grant No. IIS-1350522. 

%
%
%
%

\bibliography{iclr2019_conference}
\bibliographystyle{iclr2019_conference}

\appendix
\section{Appendix}
\subsection{Dataset Descriptions}
\label{dataset}
\subsubsection{MIMIC-III Dataset}
We evaluate our model framework on the publicly available MIMIC-III dataset \citep{johnson2016mimic}. MIMIC-III is a de-identified
dataset collected at Beth Israel Deaconess Medical Center
from 2001 to 2012. It consists of approximately 58,000 hospital admission records.  This data set contains sparse and irregularly sampled physiological signals, medications, diagnostic codes, in-hospital mortality, length of stay and more. We focus on predicting in-hospital mortality and length of stay using the first 48 hours of data. We extracted 12 standard physiological variables from each of the 53,211 records obtained after removing hospital admission records with length of stay less than 48 hours. Table \ref{table:mis} shows the features, sampling rates (per hour) and their missingness information computed using the union of all time stamps that exist in any dimension of the input time series. 

\begin{table}[h]
\centering
\caption{Features extracted from MIMIC III for our experiments}
\label{table:mis}
                \begin{tabular}[h]{l c c}
                
                 \toprule
                 {feature} & {\#Missing} & {Sampling Rate}\\
                    \midrule
                    SpO2 & $31.35\%$ & $0.80$\\
                    HR  & $23.23\%$ & $0.90$\\
                    RR & $59.48\%$  & $0.48$\\
                    SBP & $49.76\%$ & $0.59$\\
                    DBP & $48.73\%$ & $0.60$\\
                    Temp & $83.80\%$& $0.19$\\
  
                    \bottomrule
                    \end{tabular}  
                    \,\,\,\,\,\,
                \begin{tabular}[h]{l c c}
                 \toprule
                 {feature} & {\#Missing} & {Sampling Rate}\\
                    \midrule
                    TGCS & $87.94\%$& $0.14$\\
                    CRR &$95.08\%$& $0.06$\\
                    UO & $82.47\%$&$0.20$\\
                    FiO2 &$94.82\%$&$0.06$\\
                    Glucose &$91.47\%$&$0.10$\\
                    pH &$96.25\%$&$0.04$\\
                    \bottomrule
                    \end{tabular}
                \end{table}

\cut{
\begin{table}[h]
\caption{Features extracted from MIMIC III for our experiments}
\label{table:mis}
            \footnotesize
                \begin{tabular}[h]{l c}
                 \toprule
                 {feature} & {\#Missing}\\
                    \midrule
                    SpO2 & $31.35\%$\\
                    HR  & $23.23\%$\\
                    RR & $59.48\%$\\
                    
                    \bottomrule
                \end{tabular}
               \hfill
                \begin{tabular}[h]{l c}
                 \toprule
                 {feature} & {\#Missing}\\
                    \midrule
                    SBP & $49.76\%$\\
                    DBP & $48.73\%$\\
                    Temp & $83.80\%$\\
  
                    \bottomrule
                    \end{tabular}
                 \hfill
                \begin{tabular}[h]{l c}
                 \toprule
                 {feature} & {\#Missing}\\
                    \midrule
                    TGCS & $87.94\%$\\
                    CRR &$95.08\%$\\
                    UO & $82.47\%$\\
                    
                    \bottomrule
                    \end{tabular}
                     \hfill
                \begin{tabular}[h]{l c}
                 \toprule
                {feature} & {\#Missing}\\
                    \midrule
                    FiO2 &$94.82\%$\\
                    Glucose &$91.47\%$\\
                    pH &$96.25\%$\\
                    \bottomrule
                    \end{tabular}
                \end{table}
}
\subsubsubsection{Prediction Tasks}

In our experiments, each admission record corresponds to one data case $(\mbf{s_n},y_n)$. Each data case $n$ consists of a sparse and irregularly sampled time series $\mbf{s}_n$ with $D=12$ dimensions. Each dimension $d$ 
of $\mbf{s}_n$ corresponds to one of the 12 vital sign time series mentioned above. In the case of classification, $y_n$ is a binary indicator where $y_n=1$ indicates that the patient died at any point within the hospital stay following the first 48 hours and $y_n=0$ indicates that the patient was discharged at any point after the first 48 hours. There are 4310 (8.1\%) patients with a $y_n=1$ mortality label. The complete data set is $\mathcal{D}=\{(\mbf{s_n},y_n)|n=1,...,N\}$, and there
are $N=53,211$ data cases. The goal in the classification task is to learn a classification function $g$ of the
form $\hat{y}_n \leftarrow g(\mbf{s}_n)$ where $\hat{y}_n$ is a discrete value.

In the case of regression, $y_n$ is a real-valued regression target corresponding to the length of stay. Since the
data set includes some very long stay durations, we let $y_n$ represent the log of the length of stay in days for 
all models. We convert back from the log number of days to the number of days when reporting results.
The complete data set is again $\mathcal{D}=\{(\mbf{s_n},y_n)|n=1,...,N\}$ with
$N=53,211$ data cases (we again require 48 hours worth of data). The goal in the regression task is to learn a regression function $g$ of the form $\hat{y}_n \leftarrow g(\mbf{s}_n)$ where $\hat{y}_n$ is a continuous value.

\subsubsection{UWave Dataset}
UWave dataset is an univariate time series data consisting of simple gesture patterns divided into eight categories. The dataset has been split into 3582  train and 896 test instances. Out of the training data, 30\% is used for validation. Each time series contains 945 observations. We follow the same data preparation method as in \cite{li2016scalable} where we randomly sample 10\% of the observations points from each time series to create a sparse and irregularly sampled data.

\subsection{Implementation Details}
\label{implementation}
\subsubsection{Proposed Model}
The model is learned using the Adam optimization method 
in TensorFlow with gradients provided via automatic differentiation.
However, the actual multivariate time series  
representation used during learning is based on the union
of all time stamps that exist in any dimension of the input time series.
Undefined observations are represented as zeros and a separate
missing data mask is used to keep track of which time series have
observations at each time point. Equations \ref{eq:interp0} to \ref{eq:interp4} are modified such
that data that are not available are not taken into account at all.
This implementation is exactly equivalent to the computations
described, but supports parallel computation
across all dimensions of the time series for a given data case.

Finally, we note that the learning problem can be solved using a
doubly stochastic gradient based on the use of mini batches
combined with re-sampling the artificial missing data masks 
used in the interpolation loss. In practice, we randomly select
$20\%$ of the observed data points to hold out from every input time
series. 

For the time series missing entirely, our interpolation network assigns the starting point (time t=0) value of the time series to the global mean before applying the two-layer interpolation network. In such cases, the first interpolation layer just outputs the global mean for that channel, but the second interpolation layer performs a more meaningful interpolation using the learned correlations from other channels.

\subsubsection{Baselines}
The Logistic Regression model is trained with cross entropy loss with regularization strength set to 1. The support vector classifier is used with a RBF kernel and trained to minimize the soft margin loss. We use the cross entropy loss on the validation set to select the optimal number of estimators in case of Adaboost and Random Forest. Similar to the classification setting, the optimal number of estimators for regression task in Adaboost and Random Forest is chosen on the basis of squared error on validation set.

\subsubsection*{MIMIC-III Dataset}
We evaluate all models using a five-fold cross-validation estimate of generalization performance.
In the classification setting,  all the deep learning baselines are trained to minimize the cross entropy loss while the proposed model uses a composite loss consisting of cross-entropy loss and interpolation loss (with $\delta_R = 1$) as described in section $3.2.3$. In the case of the regression task, all baseline models are trained to minimize squared error and the proposed model is again trained with a composite loss consisting of squared error and interpolation loss.

We follow the multi-task Gaussian process implementation given by \cite{futoma2017improved} and treat the number of hidden units and hidden layers as hyper-parameters.  For all of the GRU-based models, we use the already specified parameters \citep{che2016recurrent}. The models are learned using the Adam optimization. Early
stopping is used on a validation set sub-sampled from the training folds. In the classification case,
the final outputs of the GRU hidden units are used in a logistic layer that predicts the class.
In the regression case, the final outputs of the GRU hidden units
are used as input for a dense hidden layer with $50$ units, followed by a linear output layer. 

\subsubsection*{UWave Dataset} 
We independently tune the hyper-parameters of each baseline method. For GRU-based methods, hidden units are searched over the range $\{2^5, 2^6 , \cdots, 2^{11}\}$. Learning is done in same way as described above.  We evaluate all the baseline models on the test set and compare the training time and accuracy. For the Gaussian process model, we use the squared exponential covariance function.  We use the same number of inducing points for both the Gaussian process and the proposed model. The Gaussian process model is jointly trained with the GRU using stochastic gradient descent with Nesterov momentum. We apply early stopping based on the validation set.

\subsection{Additional Experiments}
\label{ablation}

In this section, we address the question of the relative information content of the different 
outputs produced by the interpolation network used in the proposed model for MIMIC-III dataset.
Recall that for each of the $D=12$ vital sign time series, the interpolation
network produces three outputs: a smooth interpolation output (SI), a non-smooth
or transient output (T), and an intensity function (I). The above results
use all three of these outputs. 

To assess the impact of each of the interpolation network outputs, we conduct a
set of ablation experiments where we consider using all sub-sets of outputs 
for both the classification task and for the regression task. 

Table \ref{table:analysis} shows the results from five-fold cross validation mortality and length of stay prediction experiments. When using each output individually, smooth 
interpolation (SI) provides the best performance in terms of classification. Interestingly,
the intensity output is the best single information source for the regression task
and provides at least slightly better mean performance than any of the baseline methods 
shown in Table \ref{table:1}. Also interesting is the fact that the transients output 
performs significantly worse when used alone than either the smooth interpolation 
or the intensity outputs in the classification task.  

\begin{table}[h]
\caption{Performance of all subsets of the interpolation network outputs
on Mortality (classification) and Length of stay prediction (regression) tasks. SI: Smooth Interpolation, I: Intensity, T: Transients, Loss: Cross-Entropy Loss, MedAE: Median Absolute Error, EV: Explained variance }
\footnotesize
\begin{center}

\begin{tabular}{ c c c c c c} 
 \toprule
 {\bf Model}&  \multicolumn{3}{c}{\bf Classification} &  \multicolumn{2}{c}{\bf Regression}\\
 \midrule
 {} & {\bf AUC} & {\bf AUPRC} & {\bf Loss} & {\bf MedAE} & {\bf EV score}\\
 \midrule
 SI, T, I & ${\bf 0.853 \pm 0.007}$ & ${\bf 0.418\pm0.022}$& ${\bf 0.210 \pm 0.004}$  & $2.862\pm0.166$ & $0.245\pm0.019$ \\
 SI, I & $0.852 \pm 0.005$ & $0.408\pm0.017$ & $0.210 \pm 0.004$  & $2.745\pm0.062$ & $0.224\pm0.010$ \\
  SI, T & $0.820 \pm 0.008$ & $0.355\pm0.024$& $0.226 \pm 0.005$ & $2.911\pm0.073$ & $0.182\pm0.009$\\
 SI & $0.816 \pm 0.009$ &$0.354\pm0.018$& $0.226 \pm 0.005$ & $3.035\pm0.063$ &	$0.183\pm0.016$\\	 
I & $0.786\pm 0.010$ & $0.250\pm0.012$& $0.241 \pm 0.003$	 &  ${\bf 2.697\pm0.072}$ & 	$0.251\pm0.009$\\	
 I, T & $0.755 \pm 0.012$ & $0.236\pm0.014$&	$0.272 \pm 0.010$ & $2.738\pm0.101$ & ${\bf 0.290\pm0.010}$\\	
 T & $0.705 \pm 0.009$ & $0.192\pm0.008$&	$0.281 \pm 0.004$ & $2.995\pm0.130$ &	$0.207\pm0.024$ \\
  \bottomrule
 \end{tabular}
\end{center}

\label{table:analysis}
\end{table}

When considering combinations of interpolation network components, we can
see that the best performance is obtained when all three outputs are
used simultaneously in classification tasks.  For the regression task, the 
intensity output provides better performance in terms of median absolute 
error while a combination of intensity and transients output provide better explained variance score.
However, the use of the transients output contributes almost
no improvement in the case of the AUC and cross entropy loss for classification
relative to using only smooth interpolation and intensity. Interestingly,
in the classification case, there is a significant boost in performance 
by combining smooth interpolation and intensity relative to using either 
output on its own. In the regression setting, smooth interpolation appears
to carry little information. 

\subsection{Benchmark MIMIC-III Dataset}
\label{benchmark}
In this section, we compare the performance of the proposed model on a previous MIMIC-III benchmark dataset \citep{benchmark}. This dataset only consists of patients with age $> 18$. Again, we focus on predicting in-hospital mortality using the first 48 hours of data. This yields training and test sets of size 17,903 and 3,236 records respectively. 

We compare the proposed model to multiple baselines from \cite{benchmark}. In all the baselines, the sparse and irregularly sampled time-series data has been discretized into 1-hour intervals. If there are multiple observations in an interval, the mean or last observation is assigned to that interval, depending on the baseline  method. Similarly, if an interval contains no observations, the  mean or forward filling approach is used to assign a value depending on the baseline method. We compare with a logistic regression model and a standard LSTM network. In the multitask setting, multiple tasks are predicted jointly. Unlike the standard LSTM network where the output/hidden-state from the last time step is used for prediction, we provide supervision to the model at each time step. In this experiment, we use an LSTM as the prediction network in the proposed model to match the baselines. 

\begin{table}[h]
\caption{Classification performance for in-hospital mortality prediction task on benchmark dataset}
\begin{center}
\begin{tabular}{ c c c} 
 \toprule
 {\bf Model} & {\bf AUC score} & {\bf AUPRC score}\\
 \midrule
 Logistic Regression & $0.8485$ & $0.4744$ \\
 LSTM & $0.8547$ & $0.4848$ \\
 LSTM + Deep Supervision & $0.8558$ & $0.4928$ \\
 Multitask LSTM & $0.8607$ & $0.4933$ \\ 
 {\bf Interpolation Network + LSTM} & ${\bf 0.8610}$ & ${\bf 0.5370}$\\
 \bottomrule
 \end{tabular}
\end{center}
\label{table:benchmark}
\end{table}

\end{document}